# Learning Arithmetic Circuits

**Daniel Lowd** and **Pedro Domingos**
Department of Computer Science and Engineering
University of Washington
Seattle, WA 98195-2350, U.S.A.
{*lowd,pedrod*}@*cs.washington.edu*

## Abstract

Graphical models are usually learned without regard to the cost of doing inference with them. As a result, even if a good model is learned, it may perform poorly at prediction, because it requires approximate inference. We propose an alternative: learning models with a score function that directly penalizes the cost of inference. Specifically, we learn arithmetic circuits with a penalty on the number of edges in the circuit (in which the cost of inference is linear). Our algorithm is equivalent to learning a Bayesian network with context-specific independence by greedily splitting conditional distributions, at each step scoring the candidates by compiling the resulting network into an arithmetic circuit, and using its size as the penalty. We show how this can be done efficiently, without compiling a circuit from scratch for each candidate. Experiments on several real-world domains show that our algorithm is able to learn tractable models with very large treewidth, and yields more accurate predictions than a standard context-specific Bayesian network learner, in far less time.

## 1 INTRODUCTION

Bayesian networks are a powerful language for probabilistic modeling, capable of compactly representing very complex dependences. Unfortunately, the compactness of the representation does not necessarily translate into efficient inference. Networks with relatively few edges per node can still require exponential inference time. As a consequence, approximate inference methods must often be used, but these can yield poor and unreliable results. If the network represents manually encoded expert knowledge, this is perhaps inevitable. But when the network is learned from data, the cost of inference can potentially be greatly reduced, without compromising accuracy, by suitably directing the learning process.

Bayesian networks can be learned using local search to maximize a likelihood or Bayesian score, with operators like edge addition, deletion and reversal (Heckerman et al., 1995). Typically, the number of parameters or edges in the network is penalized to avoid overfitting, but this is only very indirectly related to the cost of inference. Two edge additions that produce the same improvement in likelihood can result in vastly difference inference costs. In this case, it seems reasonable to prefer the edge yielding the lowest inference cost. In this paper, we propose a learning method that accomplishes this, by directly penalizing the cost of inference in the score function.

Our method takes advantage of recent advances in exact inference by compilation to arithmetic circuits (Darwiche, 2003). An arithmetic circuit is a representation of a Bayesian network capable of answering arbitrary marginal and conditional queries, with the property that the cost of inference is linear in the size of the circuit. When context-specific independences are present, arithmetic circuits can be much more compact than the corresponding junction trees. We take advantage of this by learning arithmetic circuits that are equivalent to Bayesian networks with context-specific independence, using likelihood plus a penalty on the circuit size as the score function. Arithmetic circuits can also take advantage of other structural properties such as deterministic dependencies and latent variables; utilizing these in addition to context-specific independence is an important item of future work.

Previous work on learning graphical models with the explicit goal of limiting the complexity of inference falls into two main classes: mixture models with polynomial-time inference (e.g.: Meila and Jordan (2000); Lowd and Domingos (2005)) and graphical models with thin junction trees (e.g.: Srebro (2000); Chechetka and Guestrin (2008)). The former are limited in the range of distributions that they can compactly represent. The latter are computationally viable (at both learning and inference time) only for very low treewidths. Our approach can flexibly and compactly learn a wide variety of models, including models with very large treewidth, while guaranteeing efficient inference, by taking advantage of the properties of arithmetic circuits.

The prior work most closely related to ours is Jaeger et al.'s (2006). Jaeger et al. define probabilistic decision graphs, a new language related to binary decision diagrams. In contrast, we use standard arithmetic circuits, and our models are equivalent to standard Bayesian networks. Jaeger et al. speculate that learning arithmetic circuits directly from data would be very difficult. In this paper we propose one approach to doing this.

The remainder of our paper is organized as follows. In Sections 2 and 3, we provide background on Bayesian networks and arithmetic circuits, respectively. We describe in detail our algorithm for learning arithmetic circuits in Section 4. Section 5 contains our empirical evaluation on three real-world datasets, and we conclude in Section 6.

## 2 BAYESIAN NETWORKS

A *Bayesian network* encodes the joint probability distribution of a set of $n$ variables, $\{X_1, \ldots, X_n\}$, as a directed acyclic graph and a set of conditional probability distributions (CPDs) (Pearl, 1988). Each node corresponds to a variable, and the CPD associated with it gives the probability of each state of the variable given every possible combination of states of its parents. The set of parents of $X_i$, denoted $\Pi_i$, is the set of nodes with an arc to $X_i$ in the graph. The structure of the network encodes the assertion that each node is conditionally independent of its non-descendants given its parents. The joint distribution of the variables is thus given by $P(X_1, \ldots, X_n) = \prod_{i=1}^{n} P(X_i | \Pi_i)$.

For discrete domains, the simplest form of CPD is a conditional probability table. When the structure of the network is known, learning reduces to estimating CPD parameters. When the structure is unknown, it can be learned by starting with an empty or prior network and greedily adding, deleting and reversing arcs to optimize some score function (Heckerman et al., 1995). The score function is usually log-likelihood plus a complexity penalty or a Bayesian score (product of prior and marginal likelihood).

The goal of inference in Bayesian networks is to answer arbitrary marginal and conditional queries (i.e., to compute the marginal distribution of a set of query variables, possibly conditioned on the values of a set of evidence variables). One common method is to construct a *junction tree* from the Bayesian network and pass messages from the leaves of this tree to the root and back. A junction tree is constructed by connecting parents of the same variable, removing arrows, and triangulating the resulting undirected graph (i.e., ensuring that all cycles of length four or more have a chord). Each node in the junction tree corresponds to a *clique* (maximal completely connected subset of variables) in the triangulated graph. Ordering cliques by the highest-ranked variable they contain, each clique is connected to a predecessor sharing the highest number of variables with it. The intersection of the variables in two adjacent cliques is called the *separator* of the two cliques. A junction tree satisfies two important properties: each variable in the Bayesian network appears in some clique with all of its parents; and if a variable appears in two cliques, it appears in all the cliques on the path between them (the *running intersection property*). The *treewidth* of a junction tree is one less than the maximum clique size. The complexity of inference is exponential in the treewidth. Finding the minimum-treewidth junction tree is NP-hard (Arnborg et al., 1987). Inference in Bayesian networks is #P-complete (Roth, 1996).

Because exact inference is intractable, approximate methods are often used, of which the most popular is *Gibbs sampling*, a form of Markov chain Monte Carlo (Gilks et al., 1996). A Gibbs sampler proceeds by sampling each non-evidence variable in turn conditioned on its Markov blanket (parents, children and parents of children). The distribution of the query variables is then approximated by computing, for each possible state of the variables, the fraction of samples in which it occurs. Gibbs sampling can be very slow to converge, and many MCMC variations have been developed, but choosing and tuning one for a given application remains a difficult, labor-intensive task. Diagnosing convergence is also difficult.

### 2.1 LOCAL STRUCTURE

Table CPDs require exponential space in the number of parents of the variable. A more scalable approach is to use *decision trees* as CPDs, taking advantage of context-specific independencies (i.e., a child variable is independent of some of its parents given some values of the others) (Boutilier et al., 1996; Friedman & Goldszmidt, 1996; Chickering et al., 1997). The algorithm we present in this paper learns arithmetic circuits that are equivalent to this type of Bayesian network.

In a decision tree CPD for variable $X_i$, each interior node is labeled with one of the parent variables, and each of its outgoing edges is labeled with a value of that variable.[1] Each leaf node is a multinomial representing the marginal distribution of $X_i$ conditioned on the parent variable values specified by its ancestor nodes and edges in the tree.

The following two definitions will be useful in describing our algorithm.

---

[1] In general, each outgoing edge can be labeled with any subset of the variable's values, as long as the sets of labels assigned to all child edges include every variable value and are disjoint with each other. For simplicity, we limit our discussion to the case in which each edge has a single label, which Chickering et al. (1997) refer to as a *complete split*. For Boolean variables, as in our experiments, all types of splits are equivalent.

**Definition 1.** *For leaf node $D$ and $k$-valued variable $X_j$, the* split $S(D, X_j)$ *replaces $D$ with $k$ new leaves, each conditioned on a particular value of $X_j$ in addition to the parent values on the path to $D$.*

**Definition 2.** *Let $D$ be a leaf from the tree CPD for $X_i$. Split $S(D, X_j)$ is valid iff $X_j$ is not a descendant of $X_i$ in the Bayesian network and no decision tree ancestor of $D$ is labeled with $X_j$.*

The first definition describes a structural update to the Bayesian network; the second one gives the conditions necessary for that update to be consistent and meaningful.

A Bayesian network can now be learned by greedily applying the best valid splits according to some criterion, such as the likelihood of the data penalized by the number of parameters. This is one version of Chickering et al.'s algorithm (1997). A number of other methods have also been proposed, such as merging leaves to obtain decision graphs (Chickering et al., 1997) or searching through Bayesian network structures and inducing decision trees conditioned on the global structure (Friedman & Goldszmidt, 1996).

## 3  ARITHMETIC CIRCUITS

The probability distribution represented by a Bayesian network can be equivalently represented by a multilinear function known as the *network polynomial* (Darwiche, 2003):

$$P(X_1 = x_1, \ldots, X_n = x_n)$$
$$= \sum_{\mathbf{X}} \prod_{i=1}^{n} I(X_i = x_i) P(X_i = x_i | \Pi_i = \pi_i)$$

where the sum ranges over all possible instantiations of the variables, $I()$ is the indicator function (1 if the argument is true, 0 otherwise), and the $P(X_i|\Pi_i)$ are the parameters of the Bayesian network. The probability of any partial instantiation of the variables can now be computed simply by setting to 1 all indicators consistent with the instantiation, and to 0 all others. This allows arbitrary marginal and conditional queries to be answered in time linear in the size of the polynomial.

Unfortunately, the size of the network polynomial is exponential in the number of variables, but it can be more compactly represented using an *arithmetic circuit*. An arithmetic circuit is a rooted, directed acyclic graph whose leaves are numeric constants or variables, and whose interior nodes are addition and multiplication operations. The value of the function for an input tuple is computed by setting the variable leaves to the corresponding values and computing the value of each node from the values of its children, starting at the leaves. In the case of the network polynomial, the leaves are the indicators and network parameters. The arithmetic circuit avoids the redundancy present in the network polynomial, and can be exponentially more compact.

Every junction tree has a corresponding arithmetic circuit, with an addition node for every instantiation of a separator, a multiplication node for every instantiation of a clique, and an addition node as the root. Thus one way to compile a Bayesian network into an arithmetic circuit is via a junction tree. However, when the network contains context-specific independences, a much more compact circuit can be obtained. Darwiche (2003) describes one way to do this, by encoding the network into a special logical form, factoring the logical form, and extracting the corresponding arithmetic circuit.

## 4  LEARNING ARITHMETIC CIRCUITS

### 4.1  SCORING AND SEARCHING

Instead of learning a Bayesian network and then compiling it into a circuit, we induce an arithmetic circuit directly from data using a score function that penalizes circuits with more edges. The score of an arithmetic circuit $C$ on an i.i.d. training sample $T$ is

$$\text{score}(C, T) = \log P(T|C) - k_e n_e(C) - k_p n_p(C)$$

where the first term is the log-likelihood of the training data, $P(T|C) = \prod_{X \in T} P(X|C)$, $k_e \geq 0$ is the per-edge penalty, $n_e(C)$ is the number of edges in the circuit, $k_p \geq 0$ is the per-parameter penalty, and $n_p(C)$ is the number of parameters in the circuit. The last two allow us to easily combine our inference-cost penalty with a more traditional one based on model complexity.

We use this formulation for simplicity; our algorithm would work equally well with a Bayesian Dirichlet score (Heckerman et al., 1995), with a prior of the form $\exp(-k_e n_e(C) - k_p n_p(C))$, since the computation of the marginal likelihood would be the same as in standard Bayesian network learning. Aside from its practical utility, a prior penalizing inference cost is meaningful if we believe the inference task being modeled can be carried out quickly, for example because humans do it. Either way, the main difficulty is that the penalty (or prior) is no longer node-decomposable, and repeatedly computing it might be very expensive. Reducing this cost is one of the key technical issues addressed in this paper.

Arithmetic circuits can be learned in the same way as Bayesian networks with local structure, by starting with an empty network and greedily applying the best splits, except that candidate structures are scored by compiling them into arithmetic circuits. However, compiling an arithmetic circuit can be computationally costly, and doing so for every candidate structure would be prohibitive. A better approach is to incrementally compile the circuit as splits are applied. Table 1 shows pseudo-code for this algorithm.

The algorithm begins by constructing the initial arithmetic

Table 1: Greedy algorithm for learning arithmetic circuits.

---

**function** LearnAC(T)
initialize circuit $C$ as product of marginals
**loop**
   $C_{best} \leftarrow C$
   **for** each valid split $S(D, V)$ **do**
     $C' \leftarrow$ SplitAC($C, S(D, V)$)
     **if** score($C', T$) > score($C_{best}, T$) **then**
        $C_{best} \leftarrow C'$
     **end if**
   **end for**
   **if** score($C_{best}, T$) > score($C, T$) **then**
     $C \leftarrow C_{best}$
   **else**
     **return** $C$
   **end if**
**end loop**

---

circuit $C$ as a product of marginal distributions:

$$C = \prod_i \sum_j I(X_i = x_{ij}) P(X_i = x_{ij})$$

This initial circuit is equivalent to a Bayesian network with no edges. In each iteration, the algorithm greedily chooses and applies the best valid split, where split validity is defined according to the equivalent Bayesian network. Each split is scored by applying it to the current circuit and counting the edges and parameters.[2] Learning ends when the algorithm reaches a local optimum, where no valid split improves the score.

### 4.2 SPLITTING DISTRIBUTIONS

The key subroutine is SplitAC, which updates an arithmetic circuit without recompiling it from scratch. Given an arithmetic circuit $C$ that is equivalent to a Bayesian network $B$ and a valid split $S(D, V)$, SplitAC returns a modified circuit $C'$ that is equivalent to $B$ after applying split $S(D, V)$. We will use the following notation to refer to distributions, parameter nodes, and indicator nodes:

$d_j$: Parameter node corresponding to the $j$th probability in the multinomial distribution $D$.

$D_i$: Leaf distribution resulting from split $S(D, V)$ that replaces $D$ when $V = i$.

$d_{ij}$: Parameter node corresponding to the $j$th probability in $D_i$.

$v_i$: Indicator node $I(V = i)$.

---

[2] All model parameters are MAP estimates, using a Dirichlet prior with all hyperparameters $\alpha_{ijk} = 1$, where $k$ ranges over the leaves of the decision tree for variable $X_i$.

Table 2: Subroutine that updates an arithmetic circuit $C$ by splitting distribution $D$ on variable $V$.

---

**function** SplitAC($C, S(D, V)$)
let $M$ be the set of mutual ancestors of $D$ and $V$
let $N$ be the set of nodes between $M$ and $V$ or $D$
**for** $i \in$ Domain($V$) **do**
   create new parameter nodes $d_{ij}$
   $N_i \leftarrow$ copy of all nodes in $N$
   **for** each $n \in N$ **do**
     let $n_i$ be the copy of $n$ in $N_i$
     **for** each child $c$ of $n$ **do**
        **if** $c = v_i$ or $c$ is inconsistent with $v_i$ **then**
           skip
        **else if** $c$ is some parameter node $d_j$ **then**
           insert edge from $n_i$ to $d_{ij}$
        **else if** $c \in N$ **then**
           let $c_i$ be the copy of $c$ in $N_i$
           insert edge from $n_i$ to $c_i$
        **else**
           insert edge from $n_i$ to $c$
        **end if**
     **end for**
   **end for**
**end for**
**for** $m \in M$ **do**
   let $n_V$ be the child of $m$ that is a $V$-ancestor
   let $n_D$ be the child of $m$ that is a $D$-ancestor
   **for** $i \in$ Domain($V$) **do**
     let $n'_V$ be the copy of $n_V$ in $N_i$
     let $n'_D$ be the copy of $n_D$ in $N_i$
     create $n_{\times_i} := v_i \times n'_V \times n'_D$
   **end for**
   create $n_+ := \sum_i n_{\times_i}$
   replace $m$'s children $n_V$ and $n_D$ with $n_+$
**end for**
delete unreachable nodes, including all $d_j$

---

Table 2 contains pseudo-code for the splitting algorithm. It might at first appear that to split $D$ on $V$ it suffices to replace references to each $d_j$ with a sum of products, $\sum_i d_{ij} v_i$. However, the resulting circuit would then be correct only when $V$ is fixed to a particular value, and summing out $V$ would produce inconsistent results. Intuitively, the circuit must maintain the running intersection property of the corresponding junction tree, so that no variable can take on different values in different subcircuits. SplitAC maintains a consistent probability distribution by preserving three properties, analogous to those defined by Darwiche (2002) for logical circuits.

**Definition 3.** *For an arithmetic circuit, $C$:*

*$C$ is smooth if, for each addition node, all children are ancestors of indicator nodes for the same variables and pa-*

*rameter nodes from distributions of the same variables.*

*C is* decomposable *if, for each multiplication node, no two children are ancestors of indicator nodes for the same variable or parameter nodes from distributions of the same variable.*

*C is* deterministic *if, for each addition node, there is a variable V such that each child is the ancestor of some non-empty set of indicator nodes for V, and their sets are disjoint.*

The network polynomial for a Bayesian network contains one term for each configuration of its variables; each term includes exactly one indicator variable and one conditional probability parameter per variable. Intuitively, if $C$ is not smooth, then some terms in the polynomial it computes may not have an indicator variable and a conditional probability parameter for every variable. If $C$ is not decomposable, then some terms in the polynomial may have more than one indicator variable or conditional probability parameter for some variable. If $C$ is not deterministic, then there may be multiple terms for the same set of indicator variables.

**Definition 4.** *We define three special types of node in the circuit as follows:*

*A D-ancestor is any leaf $d_j$ corresponding to a parameter of D, or any parent of a D-ancestor.*

*A V-ancestor is any leaf $v_i$ corresponding to an indicator of V, or any parent of a V-ancestor.*

*A mutual ancestor (MA) of D and V is a node that is both a D-ancestor and a V-ancestor, and has no child that is both a D-ancestor and a V-ancestor.*

Note that every MA must be a multiplication node, or the circuit would not be smooth. Furthermore, from decomposability, each MA must have exactly one $D$-ancestor child, $n_D$, and one $V$-ancestor child, $n_V$. Naively replacing $d_j$ with $\sum_i d_{ij} v_i$ would cause both $n_V$ and $n_D$ to be ancestors of $v_i$, violating decomposability.

To avoid this, SplitAC duplicates the subcircuits between the MAs and the parameter nodes $d_j$, and between the MAs and the indicator nodes $v_i$, "conditioning" each copy on a different value of $V$. Each $n_V$ and $n_D$ are replaced by a new addition node, $n_+$, that sums over products of $v_i$ and copies of $n_V$ and $n_D$ conditioned on $v_i$. This duplication of subcircuits is the reason different splits can have widely different edge costs. We now describe the details of which nodes are duplicated and how they are connected.

Let $N$ be the set of all $D$-ancestors and $V$-ancestors that are also descendants of a mutual ancestor. These are all the nodes "in between" $D$ and $V$ that must agree on the value of $V$. For each value $i$ in the domain of $V$, SplitAC creates a copy $N_i$ of the nodes in $N$.

Let $n_i \in N_i$ be the copy of node $n \in N$. SplitAC inserts edges from $n_i$ to its children as follows. If $n$ has a child $c \in N$, then it inserts an edge from $n_i$ to the corresponding copy $c_i$. If $n$ has a child $c \notin N$, then it inserts an edge from $n_i$ to $c$. This minimizes node duplication by linking to existing nodes or copies whenever possible.

A few additional changes are required for $N_i$ to properly depend on $v_i$. If $n_i \in N_i$ has some parameter node $d_j$ as a child, SplitAC replaces it with $d_{ij}$. This is how the new leaf distributions, conditioned on $V$, are integrated into the circuit. Secondly, if $n_i$ has $v_i$ as a child, it should be omitted: every node in $N_i$ will depend on $v_i$, so this is redundant. Finally, if $n_i$ has a child that is an ancestor of some $v_j$ but not of $v_i$, then that child is inconsistent with conditioning on $v_i$ and must be removed.

Finally, SplitAC connects each mutual ancestor, $m$, to a sum over these copies. SplitAC removes the $D$-ancestor, $n_D$, and the $V$-ancestor, $n_V$, as children of $m$ and replaces them with an addition node with one child for each value of $V$. The $i$th child of the addition node is a product of $v_i$, the copy of $n_D$ from $N_i$, and the copy of $n_V$ from $N_i$. (If $m$ was an ancestor of only certain values of $V$, the addition node sums only over those values.)

Intuitively, the resulting circuit represents the correct probability distribution because $D$ has been replaced with the split distributions $D_i$, each conditioned on $v_i$, and because the circuit satisfies the running intersection property, since all nodes between $V$ and $D$ now depend on $V$.

**Theorem 1.** *After each iteration of LearnAC, C computes the network polynomial of a Bayesian network constructed by starting with an empty network and applying the same splits that were applied to C up to that iteration.*

A proof sketch is in the appendix; a complete proof can be found in (Lowd & Domingos, 2008).

### 4.3 OPTIMIZATIONS

We now discuss optimizations necessary to make this algorithm practical for real-world datasets with many variables.

Consider once again the high-level overview in Table 1. Scoring every possible circuit in every iteration would be very expensive. Choosing the split that leads to the best scoring circuit is equivalent to choosing the split that leads to the greatest increase in score, so we can store changes in score instead. The improvement in log-likelihood is not affected by other splits, and so this only needs to be computed once for each potential split. Unfortunately, the number of edges that a split adds to the circuit can increase or decrease due to other splits. For convenience, we will refer to the number of edges added by the application of a split as its *edge cost*.

As a simple example, consider a chain-structured junction tree of 5 variables: AB-BC-CD-DE-EF. If we add an arc from A to F, then A is added to every other cluster: AB-ABC-ACD-ADE-AEF. However, this also reduces the cost of adding an arc from A to E, since the two variables now appear together in a cluster. As a second example, suppose that we instead added an arc from B to F: AB-BC-BCD-BDE-BEF. Now the cost of adding an arc from A to F is greatly increased, since adding a variable to a larger cluster costs more edges than adding a variable to a smaller cluster.

Evaluating the edge cost of every potential split in every iteration is expensive. The number of potential splits is linear in the number of splits that have been performed so far, leading to a time complexity that is at least quadratic in the total number of splits. Further, computing the edge cost for a single candidate may be linear in the size of the current circuit. With a non-zero edge cost, circuit size tends to be linear in the number of iterations, leading to an $O(n^3)$ algorithm. While this is still polynomial, it makes learning models with thousands of splits intractable in practice.

Fortunately, most splits only change a fraction of edge costs. Determining exactly which costs need to be updated is difficult, but we can rule out many splits whose costs do not need to be updated using the following conservative rule. Applying one split may change the edge cost of another split $S(D, V)$ if the applied split changes a node that is an ancestor of $D$ and not $V$, or of $V$ and not $D$. This covers all nodes that lie between $D$ or $V$ and their mutual ancestors, and thus all nodes that are copied by the splitting procedure. An applied split changes a node when it copies that node or reduces the number of children it has. In practice, this single heuristic lets us avoid recomputing over 95% of the edge costs.

As an alternative to this optimization, we have found a heuristic that leads to even larger speed-ups, but at the cost of no longer being perfectly greedy. We noticed that when edge costs changed, they rarely decreased. If a split's last computed edge cost was always a valid lower bound on the true value, then we could ignore any split whose total estimated score was worse than the best split found so far in this iteration. This assumption is often not valid in practice, but it lets us learn models that are nearly as effective in an order of magnitude less time.

Two other optimizations combine well with either of the above to offer further gains. First, we can reduce the number of computations by placing potential splits in order of decreasing likelihood gain, so that we consider the splits with the highest possible scores first. Since the likelihood gain is an upper bound on the score gain, once the score of the best split found so far is greater than the next likelihood gain, this split is guaranteed to be the highest-scoring one overall.

Second, we can exit the edge calculation procedure once we know that the edge cost is sufficient to make the overall score negative. It is also possible to exit once we know that the score of the current split will be worse than the best split so far, but this interferes with the other optimizations. If we only compute an upper bound on the score, we will often have to recompute the edge cost when the next iteration requires a slightly lower upper bound.

## 5 EXPERIMENTS

### 5.1 DATASETS

We evaluated our methods on three widely used real-world datasets. The KDD Cup 2000 clickstream prediction dataset (Kohavi et al., 2000) consists of web session data taken from an online retailer. Using the subset of Hulten and Domingos (2002), each example consists of 65 Boolean variables, corresponding to whether or not a particular session visited a web page matching a certain category. Anonymous MSWeb is visit data for 294 areas (Vroots) of the Microsoft web site, collected during one week in February 1998. It can be found in the UCI machine learning repository (Blake & Merz, 2000). EachMovie[3] is a collaborative filtering dataset in which users rate movies they have seen. We took a 10% sample of the original dataset, focused on the 500 most-rated movies, and reduced each variable to "rated" or "not rated". For KDD Cup and MSWeb, we used the training and test partitions provided with the datasets. For EachMovie, we randomly selected 10% of the data for the test set and used the remainder for training.

Table 3: Summary of experimental datasets.

| Domain | Vars. | Train Exs. | Test Exs. | Density |
| --- | --- | --- | --- | --- |
| KDD Cup | 65 | 199,999 | 34,955 | 0.0079 |
| MSWeb | 294 | 32,711 | 5,000 | 0.0102 |
| EachMovie | 500 | 6,117 | 591 | 0.0581 |

Basic statistics for each dataset are shown in Table 3. Density refers to the fraction of non-zero entries across all examples and all variables.

### 5.2 LEARNING

For each dataset, we randomly split the training data into tuning and validation sets, corresponding to 90% and 10% of the training data, respectively. All parameters were tuned by training models on the tuning data and selecting the parameter sets that led to the highest log likelihood of the validation set. Finally, models were retrained using the full training set. All experiments were run on CPUs with 4 GB of RAM running at 2.8 GHz.

---
[3] Provided by Compaq at http://research.compaq.com/SRC/eachmovie/; no longer available for download, as of October 2004.

We used two versions of the algorithm for learning arithmetic circuits from Section 4: AC-Greedy, which guarantees that we pick the best split in each iteration, and AC-Quick, which uses a heuristic to avoid recomputing edge costs but may sometimes choose worse splits. We varied the per-edge cost $k_e$ from 1.0 to 0.01. Not surprisingly, our models were most accurate on the validation set with low per-edge costs (0.01 or 0.02). We also tuned the per-parameter cost $k_p$. For KDD Cup, the best cost was 0.0; for MSWeb and EachMovie, the best costs were 1.0 for greedy ACs and 0.5 for quick ACs.

We used the WinMine Toolkit (Chickering, 2002) as a baseline. WinMine implements the algorithm for learning Bayesian networks with local structure described in Section 2 (Chickering et al., 1997), and has a number of other state-of-the-art features. We tuned WinMine's multiplicative per-parameter penalty $\kappa$; the best values were: 1 (no penalty) for KDD Cup, 0.1 for MSWeb, and 0.01 for EachMovie. We looked into using thin junction trees as a second baseline, but they do not scale to datasets of these dimensions.

A summary of the learned models appears in Table 4. For each dataset, we report the log-likelihood per example on the test data, the number of edges in the arithmetic circuit, the number of leaves across all decision trees, the average and maximum number of parents across all variables, the treewidth (estimated using a min-fill heuristic), the number of edges generated by compiling the Bayesian network using c2d[4], and the training time. On each model for which c2d ran out of memory, we obtained a lower bound by compiling a model with fewer splits, obtained by halting the learning process early. We varied the number of splits until we found the most complex sub-model that could still be compiled, within 10 splits. For WinMine, the chosen sub-models had less than one quarter of the original splits.

The test-set log-likelihoods of the AC learners and WinMine are very similar, with WinMine having a slight edge. This is not surprising, given that WinMine is free to choose expensive splits. Perhaps more remarkable is that this freedom translates to very little improvement in likelihood. The difference in accuracy between quick and greedy ACs is negligible except in the case of EachMovie, where the greedy AC is actually less accurate because it did not converge in the allowed time (72h).

Not surprisingly, WinMine is much faster than the AC learners. It is worth noting that the cost of learning is only incurred once, while the cost of inference is incurred many times. Also, the AC learner directly outputs an arithmetic circuit, while WinMine's Bayesian network would still have to be compiled into one, which can be very time-consuming. Finally, the quick heuristic offers up to an order-of-magnitude speedup with similar accuracy; additional heuristics might offer additional improvements.

---

[4]Available at http://reasoning.cs.ucla.edu/c2d/. We also tried using the ACE package, but it does not support decision tree CPDs and, for our models, tabular CPDs would be prohibitively large.

Table 4: Summary of Learned Models

| KDD Cup | AC-Greedy | AC-Quick | WinMine |
|---|---|---|---|
| Log-likelih. | −2.16 | −2.16 | −2.16 |
| Edges | 382K | 365K | |
| Leaves | 4574 | 4463 | 2267 |
| Avg. parents | 13.2 | 13.0 | 16.3 |
| Max. parents | 37 | 36 | 35 |
| Treewidth | 38 | 38 | 53 |
| c2d edges | >18.2M | 3664k | >39.5M |
| Time | 50h | 3h | 3m |

| MSWeb | AC-Greedy | AC-Quick | WinMine |
|---|---|---|---|
| Log-likelih. | −9.85 | −9.85 | −9.69 |
| Edges | 204K | 256K | |
| Leaves | 1353 | 1870 | 1710 |
| Avg. parents | 2.5 | 3.1 | 5.2 |
| Max. parents | 114 | 127 | 94 |
| Treewidth | 114 | 127 | 118 |
| c2d edges | >23.5M | >44.6M | >63.5M |
| Time | 8h | 3h | 2m |

| EachMovie | AC-Greedy | AC-Quick | WinMine |
|---|---|---|---|
| Log-likelih. | −55.7 | −54.9 | −53.7 |
| Edges | 155K | 372K | |
| Leaves | 4070 | 6521 | 4830 |
| Avg. parents | 5.0 | 6.5 | 8.0 |
| Max. parents | 13 | 17 | 27 |
| Treewidth | 35 | 54 | 281 |
| c2d edges | 207k | 855k | >27.3M |
| Time | >72h[5] | 22h | 3m |

### 5.3 INFERENCE

For each dataset, we used the test data to generate queries with varied numbers of randomly selected query and evidence variables. Each query asked the probability of the configuration of the query variables in the test example conditioned on the configuration of the evidence variables in the same test example.

We estimate inference accuracy as the mean log probability of the test examples's configuration across all test examples. This is an approximation (up to an additive constant) of the Kullback-Leibler divergence between the inferred distribution and the true one, estimated using the test samples. For KDD Cup and MSWeb, we generated queries from 1000 test examples; for EachMovie, we gen-erated queries from all 593 test examples.

---

[5]AC-Greedy did not finish running in the maximum allowed time of 72h. As a result, it has fewer edges and lower log-likelihood than AC-Quick.

Table 5: Average inference time per query.

| Algorithm | KDD Cup | MSWeb | EachMovie |
|---|---|---|---|
| AC-Greedy | 194ms | 91ms | 62ms |
| AC-Quick | 198ms | 115ms | 162ms |
| Gibbs-Fast | 1.46s | 1.89s | 7.22s |
| Gibbs-Medium | 11.3s | 15.6s | 42.5s |
| Gibbs-Slow | 106s | 154s | 452s |
| Gibbs-VerySlow | 1124s | 1556s | 3912s |

For the arithmetic circuits, we used exact inference. For the Bayesian networks learned using WinMine, we used Gibbs sampling. We initialized the sampler to a random state, ran it for a burn-in period, and then collected samples to estimate the probability of the queried marginal or conditional event. All estimates were smoothed by uniformly distributing a count of 1 across all states of the query variables. Since convergence is difficult to diagnose and may take prohibitively long, we ran Gibbs sampling in four scenarios: fast (one chain, 100 burn-in iterations, 1000 sampling iterations); medium (ten chains, 100 burn-in iterations, 1000 sampling iterations); slow (ten chains, 1000 burn-in iterations, 10,000 sampling iterations); and very slow (ten chains, 10,000 burn-in iterations, 100,000 sampling iterations).

Figure 1 shows the relative accuracy of the different methods on each dataset. Per-variable query log-likelihood is on the $y$ axis. In the graphs on the left, each query included 30% of the variables in the domain, conditioned on 0% to 50% of the domain variables as evidence. In the graphs on the right, the number of query variables varies from 10% to 50%, conditioned on 30% of the variables in the domain as evidence. Inference times (averaged over all queries) are listed in Table 5. Note that AC inference times are in milliseconds, while Gibbs inference times are in seconds.

The ACs were roughly one order of magnitude faster than the fastest runs of Gibbs sampling, and four orders of magnitude faster than the slowest. Except when the number of query variables is very small, the ACs also easily dominate even the slowest runs of Gibbs sampling on accuracy. Because of the approximate inference, the slightly higher test-set log-likelihood of WinMine's models does not translate into higher accuracy in answering queries. Presumably, given enough time Gibbs sampling will eventually catch up with the ACs in accuracy, but by then it will be many orders of magnitude slower. Further, Gibbs sampling (like other approximate inference methods) requires tuning for best results, and we can never be sure that it has converged. In contrast, the AC inference is reliable, the time it takes is predetermined, and the time is short enough for online or interactive use.

## 6 CONCLUSION

In the past, work on learning and inference in graphical models has been largely separate. This has had the somewhat paradoxical result that much effort is often expended to learn accurate models, only to result in less accurate predictions when approximate inference becomes necessary. Our work seeks to ameliorate this by more closely integrating learning and inference. In particular, we presented an algorithm for learning arithmetic circuits by maximizing likelihood with a penalty on circuit size. This ensures efficient inference while still providing great modeling flexibility. In experiments on real-world domains, our algorithm outperformed standard Bayesian network learning on both accuracy of query answers and speed of inference.

Directions for future work include: investigating other algorithms for learning arithmetic circuits; extending our approach to handle learning with missing data and hidden variables; applying it to Markov networks, continuous domains, and relational representations; etc.

**Acknowledgements**

The authors wish to thank Mark Chavira, Adnan Darwiche, and Knot Pipatsrisawat for help applying c2d to our Bayesian networks. This research was partly funded by a Microsoft Research fellowship awarded to the first author, DARPA contracts NBCH-D030010/02-000225, FA8750-07-D-0185, and HR0011-07-C-0060, DARPA grant FA8750-05-2-0283, NSF grant IIS-0534881, and ONR grant N-00014-05-1-0313. The views and conclusions contained in this document are those of the authors and should not be interpreted as necessarily representing the official policies, either expressed or implied, of DARPA, NSF, ONR, or the United States Government.

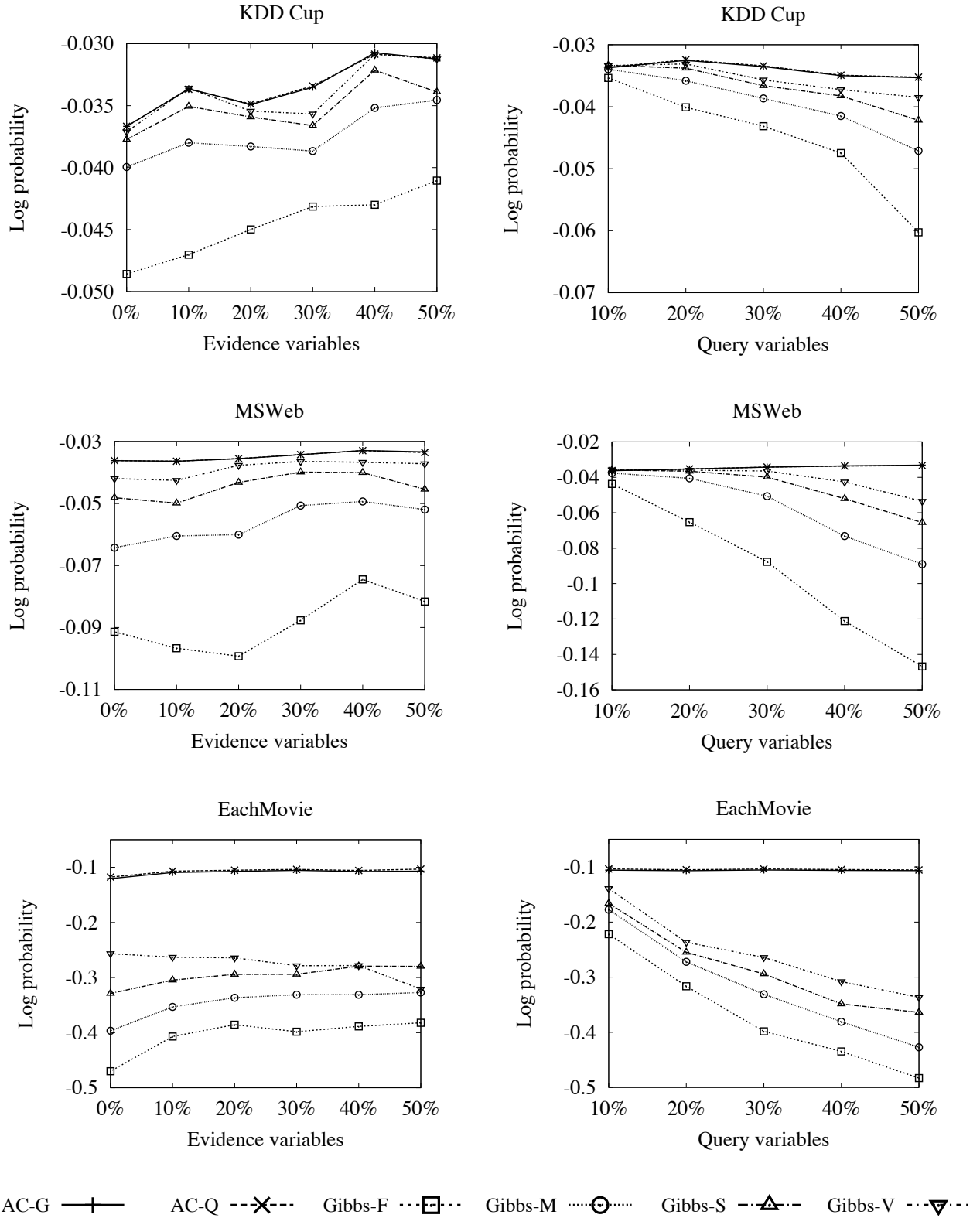

Figure 1: Conditional log probability per query variable, per query. In the legend, AC-G refers to AC-Greedy and AC-Q refers to AC-Quick. Gibbs-F, Gibbs-M, Gibbs-S and Gibbs-V refer to the fast, medium, slow, and very slow Gibbs sampling scenarios, respectively.

## APPENDIX: PROOF SKETCH FOR THEOREM 1

**Lemma 2.** *At every iteration of LearnAC, $C$ is smooth, decomposable, and deterministic.*

This can be proved by induction on the calls to SplitAC in each iteration. It is easy to verify that the initial circuit is smooth, decomposable, and deterministic. Verifying that these properties are preserved by each call to SplitAC involves a second induction over the structure of the circuit, working up from the leaf nodes. The proof can be found in Lowd and Domingos (2008).

*Proof Sketch for Theorem 1.* (By induction over the number of splits performed.) The initial circuit is a product of marginal distributions, equivalent to a Bayesian network with no arcs, so the base case is satisfied.

Assuming the circuit $C$ was equivalent to a Bayesian network $B$ after the last iteration of LearnAC, we must demonstrate that, after applying split $S(D, V)$, the resulting circuit $C'$ is equivalent to $B$ with the split $S(D, V)$.

For an arithmetic circuit, $C$, we can construct the *logical image* of $C$ by replacing addition with disjunction and multiplication with conjunction. In order to make the different values of each variable mutually exclusive, we replace indicator nodes $v_i$ with conjunctions of $v_i$ and the negation of every other $v_j$ for $j \neq i$. We apply an analogous transformation to the conditional probability parameters for each variable.

It is straightforward to show that if $C$ is a smooth, decomposable, and deterministic AC, then its logical image satisfies the equivalent properties of a logical circuit, as defined by Darwiche (2002).

Let $L$ be the logical image of $C$ and $L'$ be the logical image of $C'$. From Lemma 2 and the discussion of logical images, we know that $C$, $C'$, $L$, and $L'$ are all smooth, deterministic, and decomposable.

It can be shown inductively that the models of $L$ are the terms of the network polynomial for $B$. It can also be shown that exactly one indicator variable $v_i$ is true for each variable $V$ in every model of $L$ and $L'$. This means that each logical circuit can be expressed as a disjunction over the values of $V$: $L = \vee_i (v_i \wedge L)$, $L' = \vee_i (v_i \wedge L')$. In every model of $(v_i \wedge L)$, a node that is an ancestor of $v_j$ and not of $v_i$ is guaranteed to be false. (This can be shown using smoothness.) We can therefore remove links in $(v_i \wedge L)$ to any such node from nodes in between MAs and $V$ without affecting the truth value of the logical circuit.

We can also simplify $(v_i \wedge L')$. For any MA in $L'$, the new addition node (disjunction in $L'$) can be replaced with its $i$th child since all other children are known to be false. The $i$th child is a conjunction of $v_i$ (already assumed to be true) and a copy of two children of the MA conditioned on $v_i$. Since conjunction is associative, we can simplify the MA by linking it directly to the children of this conjunction rather than to the conjunction.

Having performed these simplifications, we can see that: $(v_i \wedge L)$ is logically equivalent to $(v_i \wedge L')$, except that parameters $d_j$ have been replaced with $d_{ij}$ in $(v_i \wedge L')$. Taking the disjunction over all $v_i$, we can conclude that the models of $L$ are identical to those of $L'$, except that whenever $d_j$ and $v_i$ are true in a model of $L$, $d_{ij}$ and $v_i$ are true in the corresponding model of $L'$. This is sufficient to demonstrate that the models of $L'$ are the terms of the network polynomial for $B$ after applying split $S(D, V)$.

Since $L'$ is smooth, deterministic, and decomposable, by Theorem 1 from Darwiche (2002), $C'$ computes the network polynomial of $B$ with the split $S(D, V)$. □